\documentclass[letterpaper, 10 pt, conference]{ieeeconf}  

\IEEEoverridecommandlockouts                              

\title{Deep Predictive Models for Collision Risk Assessment in Autonomous Driving}

\author{Mark Strickland$^{1}$, Georgios Fainekos$^{1}$, Heni Ben Amor$^{1}$ \thanks{$^{1}$Mark Strickland, Georgios Fainekos and Heni Ben Amor are with the School of Computing, Informatics and  Decision  Systems  Engineering,  Arizona  State  University,  660 S. Mill Ave, Tempe, AZ 85281 {\tt\small \{mestric1, fainekos, hbenamor\} at asu.edu} }%
}

\usepackage{graphicx}
\graphicspath{ {images/} }

\usepackage{amsmath}
\usepackage{cite}

\begin{document}

\maketitle
\thispagestyle{empty}
\pagestyle{empty}

\begin{abstract}
In this paper, we investigate a predictive approach for collision risk assessment in autonomous and assisted driving. A deep predictive model is trained to anticipate imminent accidents from traditional video streams. In particular, the model learns to identify cues in RGB images that are predictive of hazardous upcoming situations. In contrast to previous work, our approach incorporates (a) temporal information during decision making, (b) multi-modal information about the environment, as well as the proprioceptive state and steering actions of the controlled vehicle, and (c) information about the uncertainty inherent to the task. To this end, we discuss Deep Predictive Models and present an implementation using a Bayesian Convolutional LSTM. Experiments in a simple simulation environment show that the approach can learn to predict impending accidents with reasonable accuracy, especially when multiple cameras are used as input sources.  
\end{abstract}

\section{INTRODUCTION}

The concept of autonomous driving has transitioned from being an unlikely vision of robotics and artificial intelligence to being a present-day reality. A plethora of innovations in sensing, mapping, controls, and reasoning have enabled intelligent transportation systems that can deal with an impressive range of environmental conditions. Yet, an important factor that will impact the adoption of this technology in our society is the inherent risk to the human driver. Autonomous cars need to constantly assess the risk of accidents and generate steering commands accordingly. Besides autonomous driving, techniques for risk assessment are also a critical element of advanced driver-assistance systems (ADAS). Many existing approaches to collision risk assessment are based on the analysis of distances to nearby road vehicles. However, such approaches neglect a variety of visual cues that foreshadow impending accidents. Human drivers often use their prior experience to judge such risks, which in turn affect their driving patterns and level of alertness. An unusual movement of a vehicle in front, for example, can already be a sufficient cue for hitting the brakes.  

%
%
\begin{figure}
    \centering
    \includegraphics[width=0.5\textwidth]{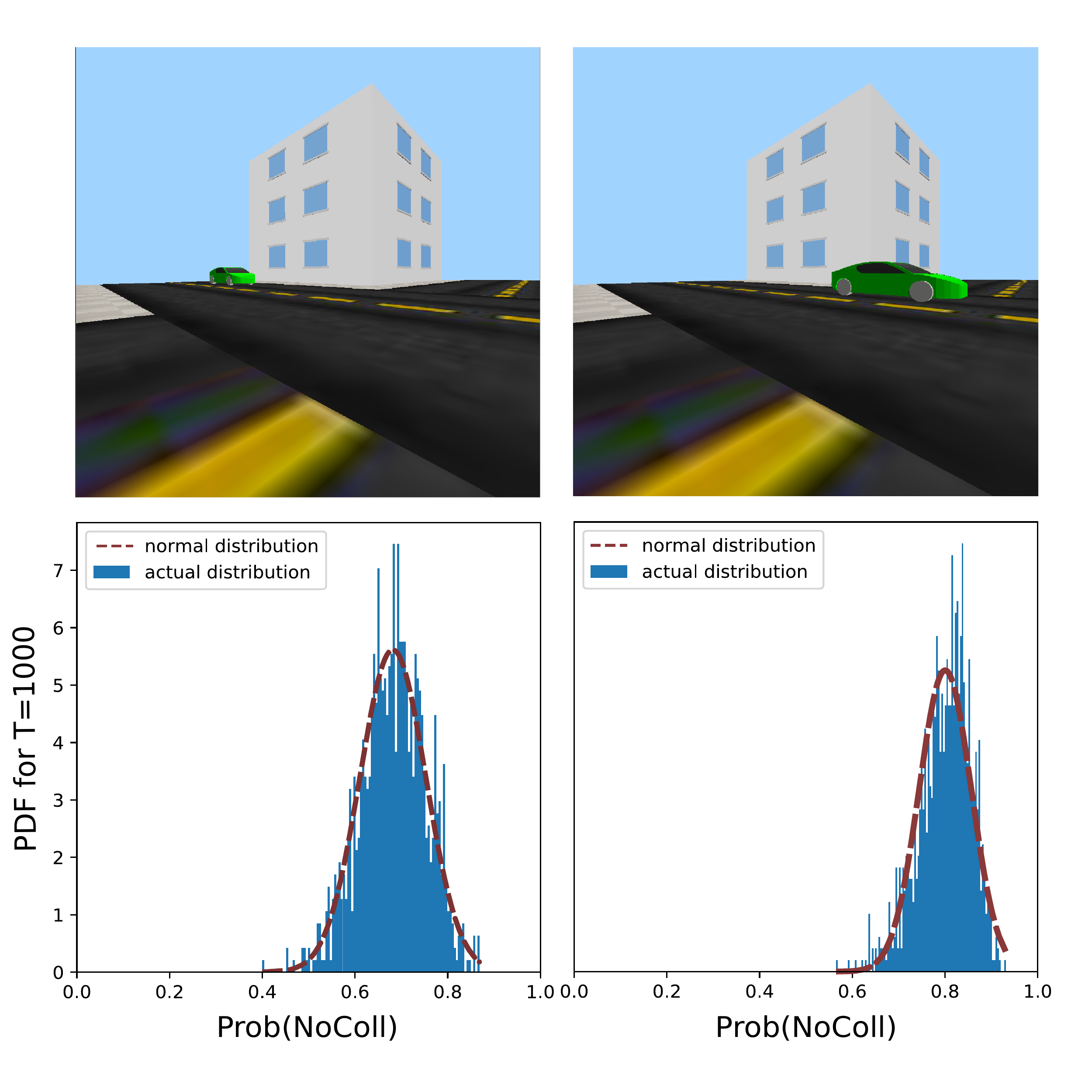}
    \caption{Collision risk assessment via Deep Learning. Recent insights on Bayesian Deep Networks are used to generate a probablity distribution. The histogram is computed out of 1000 stochastic forward passes.}
    \label{fig:hist-1k}
\end{figure}

In this paper, we investigate a predictive approach for collision risk assessment in autonomous and assisted driving. A deep predictive model is trained to anticipate imminent accidents from traditional video streams. In particular, the model learns to identify cues in RGB images that are predictive of hazardous upcoming situations. In contrast to previous work, our approach incorporates (a) temporal information during decision making, (b) multi-modal information about the environment, as well as the proprioceptive state and steering actions of the controlled vehicle, and (c) information about the uncertainty inherent to the task. The proposed approach is analyzed in a simple virtual environment under different training and test conditions. The goal of these evaluations is not to identify the best possible prediction rates. Rather, our main objective is to produce first insights on the impact of sensor placement, the importance of multi-modal input variables and network structure, as well as the nature of probabilistic outputs in this domain. In particular, we are interested in the following questions:
\begin{itemize}
\item Is there an advantage to using multiple cameras simultaneously?
\item Does the inclusion of proprioceptive data improve prediction quality?
\item Does prediction uncertainty reveal information about the current situation?
\end{itemize}

The remainder of the paper is organized as follows. First, we will discuss related work and then turn towards presenting the methodology including the Deep Predictive Model and Bayesian Convolutional LSTM at the center of our approach. Finally, we will present results on a simulated intersection scenario.

%
%
\section{RELATED WORK}
\begin{figure*} [ht!]
    \centering
    \includegraphics[width=0.9\textwidth]{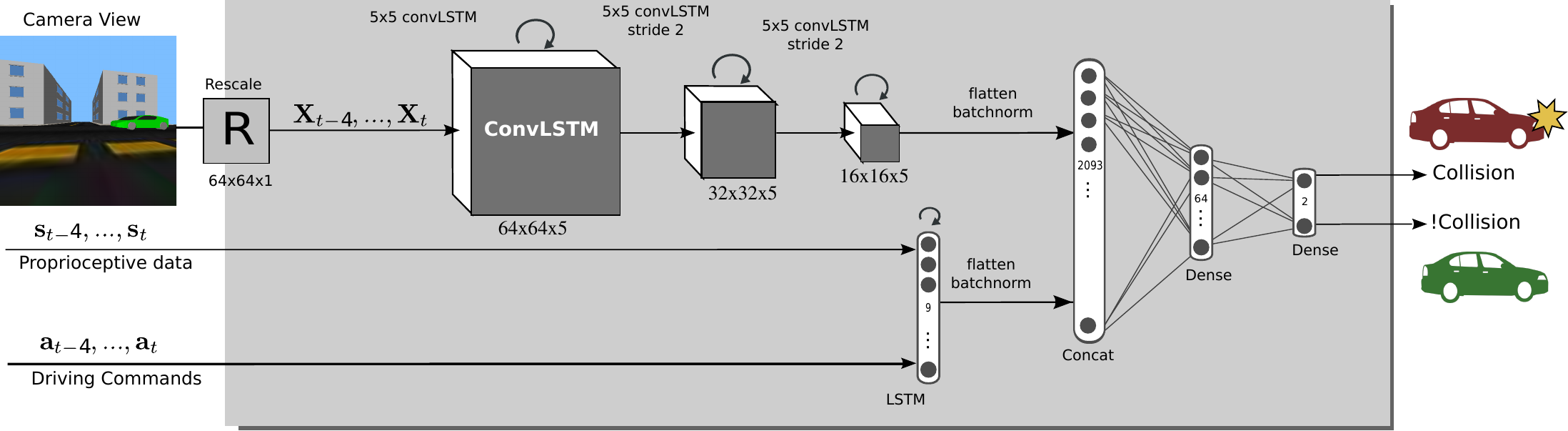}
    \caption{Architecture of the Deep Predictive Model.}
    \label{fig:arch}
\end{figure*}
Given the appropriate dataset resources, research in deep predictive models for autonomous vehicles has burgeoned in recent years.  A range of commercial interests have been working on developing autonomous vehicle technology.  See, e.g., UC Berkeley's DeepDrive\cite{Deepdrive}.  The problem of predicting impending collisions before they occur has been a popular focal point for research in this space\cite{chen2017rear, DBLP:journals/corr/HamletC15, Powar, chang2010intelligent}.  The use of deep predictive models in autonomous vehicles has been developing steam recently\cite{DBLP:journals/corr/abs-1202-2160,DBLP:journals/corr/DrewsWGTR17,hadsell2009learning}.  For example, some groups have used deep networks for lane and vehicle detection in highway driving scenarios~\cite{DBLP:journals/corr/HuvalWTKSPARMCM15}.  However, the combination of image data with multi-modal proprioceptive data for input to deep networks in a self-supervised setting has been less popular.  See, e.g.,~\cite{cho2014multi}.  Recently, Jain et al. have proposed a sensory-fusion deep learning architecture that combines GPS and street map data with image data from external sensors as well as a face camera focused on the driver, for the purpose of anticipating risky actions by the driver of the sensor car~\cite{DBLP:journals/corr/JainKSRSS16}.  However, this work employs recurrent networks only and, hence, requires feature extraction and pre-processing for the image sensor data~\cite{DBLP:journals/corr/JainKSRSS16}.  Others have also employed Bayesian inference techniques to determine model uncertainty and used the results to improve understanding of driving scenarios~\cite{DBLP:journals/corr/KendallBC15} and reduce SLAM relocalization errors~\cite{kendall2016modelling}.  However, these works do not apply  Bayesian inference to convolutional-recurrent networks.  
Here, we explore this sub-field from the perspective of developing insights regarding image sensor placement, input data strategies, and deep network architecture. In doing so, we propose a network architecture that is able to learn visuo-temporal patterns without requiring any image pre-processing. An important advantage of this approach is that the Deep Predictive Network can learn its own, task-specific filters that improve prediction performance.

Several players in the autonomous vehicle space have been working on the problem of compiling a dataset of an appropriate size for training deep neural networks for use by vehicles.  For example, Tesla plans to collect data from its vehicles already on the road having upgraded Autopilot systems~\cite{Tesla}.  Others have recognized the problem and are attempting to address it through a variety of means, including sample datasets\cite{Mtyai}, free-for-academic-use datasets\cite{Cityscapes}, and noncommercial-only datasets\cite{Geiger2013IJRR}.  Although it remains to be seen whether all (or even most) will make their datasets available to the public, at least some significant players appear amenable to pooling assets with the community~\cite{Udacity}.  Assuming others, including automakers, will follow suit, it seems clear that very large datasets including image, state, and action data will be available soon.

\section{APPROACH}
We propose a Deep Predictive Model whose primary goal is to improve vehicle occupant safety by predicting future vehicle collisions in time to activate driver warning systems.  With this goal in mind, the DPM is designed to recognize and anticipate dynamic catastrophic events beyond the immediate time horizon (e.g., collisions with another vehicle), as opposed to conventional warning and avoidance systems (e.g., backup cameras, blind-spot warning systems) which are, for the most part, designed to respond to static conditions in the immediate vicinity of the vehicle.  To perform this safety function, visual and proprioceptive data inputs are processed by a deep neural network that is trained to recognize the conditions of an impending vehicle collision, far before it occurs. 

\subsection{Deep Predictive Network}
Figure~\ref{fig:arch} depicts the basic structure of our DPM. Sequences of image sensor data (e.g., a plurality of individual images that together make up a short video), proprioceptive vehicle state data (e.g., vehicle positions, camera positions, speed), and driver action commands (e.g., commanded speed, acceleration) are fed as inputs into the network. In turn, the network generates predictions about future, upcoming collisions. As shown in Figure \ref{fig:arch}, the core of the model is a deep network containing a combination of convolutional and recurrent layers that process the input image sensor data in both space and time. This novel deep network architecture will be referred to as Bayesian ConvLSTM and will be explained in more detail below. An important feature in this regard is the ability of the DPM to work with multimodal data representing different types of available information.  



More specifically, Figure \ref{fig:arch} shows separate branches within the network which allow for multimodal input: one branch for processing image sensor data and a second branch for processing proprioceptive (sensor vehicle) state data, as well as action data.  In the case of multi-perspective image sensor data, four branches are used: three for processing image sensor data and the fourth branch processes proprioceptive data. The result is a powerful framework that extracts information from multiple data sources in both space and time.

The input to each branch is a sequence of data, including data for a plurality of time steps. The length of the sequence is a configurable parameter, but for purposes of this study we used a relatively short sequence length of five frames. Thus, each branch of the deep network takes as input a plurality of images or values of each input state/action datum, corresponding to data collected at each time step over the length of the sequence.  Proceeding through the deep network, data collected periodically over the sequence length yields sufficient input data to generate a prediction output. 

All recurrent layers transforming state/action inputs in the second branch of the deep predictive model are implemented as Long Short Term Memory (LSTM) nodes~\cite{hochreiter1997long}.  These nodes utilize purpose-built memory cells to store information, permitting processing of the inputs in time, as noted in~\cite{DBLP:journals/corr/Graves13}. The convolutional recurrent layers transforming image inputs in the first branch of the DPM are implemented as convolutional LSTM cells or nodes. Convolutional LSTM cells process the image inputs spatiotemporally, or compute the output of the cell according to the inputs and past states of its local neighbors~\cite{DBLP:journals/corr/ShiCWYWW15}.

\subsection{Bayesian ConvLSTM}
A critical feature of our approach involves estimating the uncertainty, or conversely the confidence, of the deep predictive model in the predictions generated.  Prior to making use of the model output, uncertainty metrics may be used to improve the quality and accuracy of the DPM predictions, or at least inform the judgment made as to classification (e.g., collision vs. no-collision).  For example, with basic confidence information accompanying key predictions, the model output may be filtered before use by advanced driver-assistance systems (ADAS) or autonomous controllers to prioritize predictions with confidence exceeding a threshold.

We leverage recent theoretical insights from~\cite{Gal2016Uncertainty} for estimating model uncertainty within the DPM.  According to ~\cite{Gal2016Uncertainty}, multiple stochastic forward passes using a stochastic version of Dropout are identical to variational inference in Gaussian processes. In order to incorporate this insight into a network capable of spatio-temporal processing, we modify the convolutional recurrent layers to `remove' (set to zero) a random selection number of inputs, outputs, and recurrent connections. The random selection is maintained across all time steps in the recurrently-processed sequence of input data, but is varied for each stochastic forward pass. This has the effect of modifying the weighting tensors (W) in the convolutional recurrent (ConvLSTM) gate equations by setting to zero a random number of elements, the random number being controlled by the Dropout rate $\text{r}_{d}$ (e.g., 0.01 for dropping 1 percent of the connections).  The modified weighting tensors are formed by elementwise multiplication with a matrix in which a dropout-rate fraction of random elements have been set to zero.  

Weighting tensors applied as convolutions (including $\mathbf{W}_{xi}$, $\mathbf{W}_{hi}$, $\mathbf{W}_{xf}$, $\mathbf{W}_{hf}$, $\mathbf{W}_{xc}$, $\mathbf{W}_{hc}$, $\mathbf{W}_{xo}$, $\mathbf{W}_{ho}$) are generally of the form $\mathbf{R}^{\text{m x n x p}}$, where m and n are convolution kernel dimensions and p is the number of convolution filters.  Weighting tensors applied as elementwise multiplicands (including $\mathbf{W}_{ci}$, $\mathbf{W}_{cf}$, $\mathbf{W}_{co}$) are of the form $\mathbf{R}^{\text{q x r x p}}$, where q and r are the number of rows and columns, respectively, of the input images, and p is the number of convolution filters.

The weighting tensors are employed in the expressions governing the convolutional recurrent layers, which are:

\begin{eqnarray}
\begin{aligned}
\label{Uncerteqns} 
\mathbf{I}_t &= \sigma(\mathbf{W}_{xi} \mathbf{X}_t + \mathbf{W}_{hi} \mathbf{H}_{t-1} + \mathbf{W}_{ci} \odot \mathbf{C}_{t-1} + \mathbf{b}_i)\text{,}\\
\mathbf{F}_t &= \sigma(\mathbf{W}_{xf} \mathbf{X}_t + \mathbf{W}_{hf} \mathbf{H}_{t-1} + \mathbf{W}_{cf} \odot \mathbf{C}_{t-1} + \mathbf{b}_f)\text{,}\nonumber\\
\mathbf{C}_t &= \mathbf{F}_t\odot \mathbf{C}_{t-1} + \mathbf{I}_t \odot \tanh(\mathbf{W}_{xc} \mathbf{X}_t + \mathbf{W}_{hc} \mathbf{H}_{t-1} + \mathbf{b}_c)\text{,}\\
\mathbf{O}_t &= \sigma(\mathbf{W}_{xo} \mathbf{X}_t + \mathbf{W}_{ho} \mathbf{H}_{t-1} + \mathbf{W}_{co} \odot \mathbf{C}_{t} + \mathbf{b}_o)\text{,}\\
\mathbf{H}_t &= \mathbf{O}_t \odot \tanh(\mathbf{C}_t)\text{.}
\end{aligned}
\end{eqnarray}

In the above equations, $\sigma$ is the sigmoid function, and $\mathbf{I}$, $\mathbf{F}$, $\mathbf{O}$, $\mathbf{X}$ and $\mathbf{H}$ denote the input gate, forget gate, output gate, cell inputs, and cell outputs, respectively, and ${C}$ denotes the hidden states.  Gate variables $\mathbf{I}_t$, $\mathbf{F}_t$, $\mathbf{O}_t$ denote 3D tensors with two spatial dimensions (e.g., q and r) and one convolution filter dimension (e.g., p).  Note that $\odot$ is the Hadamard product~\cite{DBLP:journals/corr/ShiCWYWW15} and all other multiplications involving tensors in the above equations are convolution operations.

We implemented MC dropout in the recurrent and convolutional-recurrent layers of the DPM.
Performing the stochastic forward passes multiple times yields a set of different outputs for the same input data. According to~\cite{Gal2016Uncertainty}, such stochastic forward passes in a deep network will approximate Variational inference in a Gaussian process. Hence, by running a number of forward passes, we can generate a histogram which approximates the distribution underlying the predictions of the neural network. 

A visualization of such stochastic forward passes can be seen in Figure~\ref{fig:hist-1k}. The top pictures of Figure~\ref{fig:hist-1k} show the current scene, while the bottom graphs show the histogram of the stochastic forward passes. The figure also shows Gaussian distributions that have been fit into the histogram. We can see that the certainty of the network about the risk of collision increases, as indicated by a higher mean value and lower variance of the distribution.

\subsection{Training Data Acquisition}

Training the above model requires visual and proprioceptive data from several time-steps, as well as a label indicating whether an accident occurred or not. All of this information is available in modern automobiles. Today, most cars are equipped with on-board cameras and a variety of sensors. The binary label indicating the occurrence of an accident is related to the activation of the airbag -- airbags deploy only in the case of strong physical perturbations, i.e., minor or major accidents. Since such a dataset is not yet available, we use simulated vehicle interactions for the generation of training data. Note that our overall goal is to draw early insights that can help the community and industry to deploy such systems in real-world scenarios. 


We create training and test data sets by varying simulation parameters including vehicle relative positioning, speeds, accelerations, sensor configuration, and driving commands. The simulation environment features a 3D representation of an intersection and two interacting cars. At each time step, the visual representation of the scene is collected from one or more `on-board' image sensors (e.g., dashcam) attached to and carried by a particular vehicle, which we designate the `sensor vehicle'.  In addition, proprioceptive data describing the state and action of the sensor vehicle, including vehicle position, speed, acceleration, image sensor position, and action data are recorded. Only data from the `sensor car' (which will later run the collision risk assessment algorithms) is recorded. No data from opposing cars is recorded. Any such data, e.g., speed of the opposing car needs to be extracted from visual cues by the DPM.

%
%
\begin{figure*}[t!]
    \centering
    \includegraphics[width=0.99\textwidth]{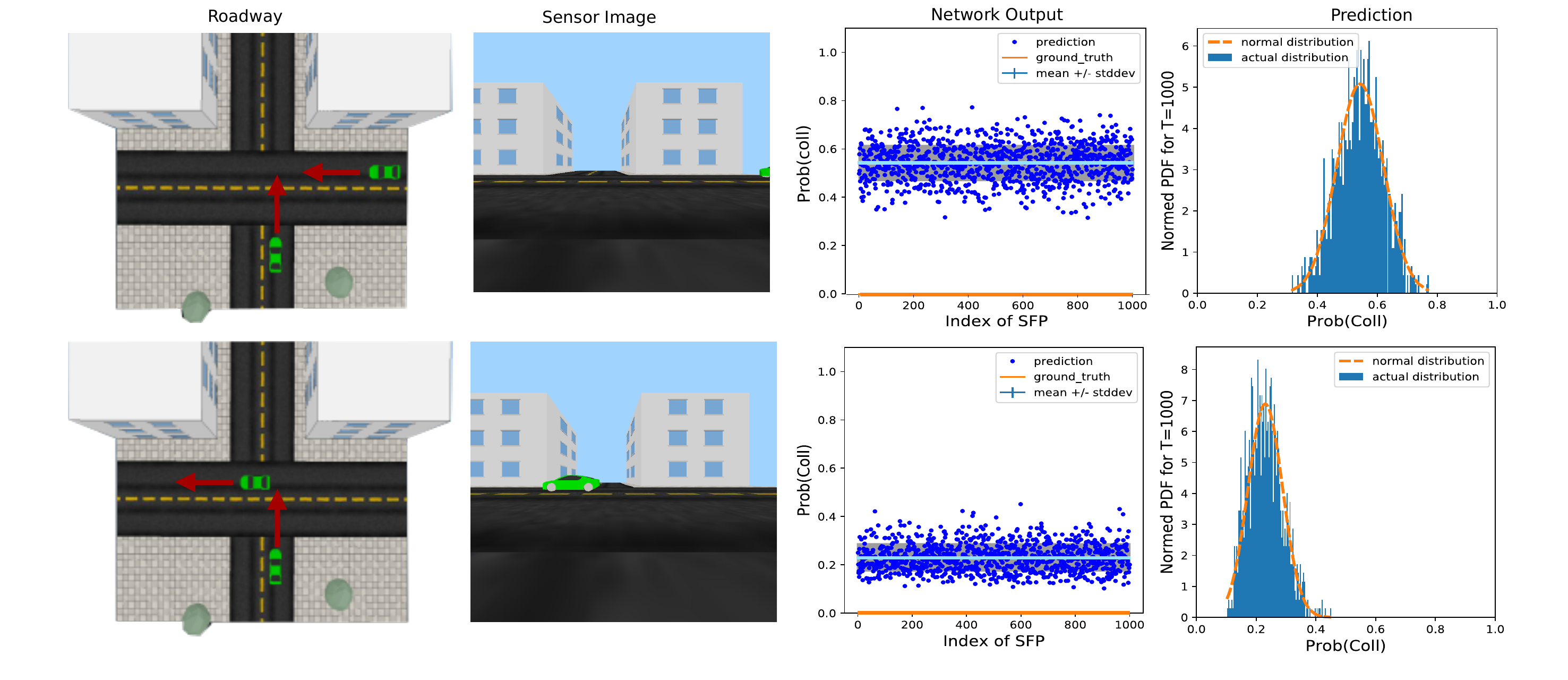}
    \caption{Birds-Eye view of roadway with sensor image and prediction.}
    \label{fig:road-sensor-bar}
\end{figure*}

We alternate positioning of the on-board image sensor at different locations on the body of the sensor car. The deep network takes as input the location of the on-board image sensor on the vehicle body along with other proprioceptive data associated with the sensor car, allowing predictions to take into account the perspective from which image data has been gathered.

\section{EXPERIMENTS AND RESULTS}
In this section, we will describe the experiments performed to validate the introduced methodology.  We first introduce the setup used for data collection and later report the results of training and validation in differing scenarios.

\subsection{Experimental Setup}    
In this study, we chose to simulate a number of vehicle interactions and vary simulation parameters in order to build our training dataset. We used a popular robotics simulation and experimentation platform~\cite{vrep2013} to simulate image sensor data, vehicle state data, and vehicle action commands.
    
    
    The training dataset was generated using a series of dynamic street scenes involving two vehicles, one of the vehicles carrying an attached image sensor, in an otherwise sparse simulated environment.  The two vehicles were pre-programmed to collide or not collide, depending on the scene and/or on a delay parameter. Sensor, state, and action data were collected at 20 Hz. We set up the sensor car so as to allow for \emph{up to three cameras} for observing the environment. This data was post-processed to assemble sequences of data, consisting of sensor, state, and action data for several time steps.  As noted, we used sequences that were five frames in length.  The sequences were assembled and randomized before being assigned to one of three categories (e.g., train, validate, test) for use in training, validating, and testing the deep predictive model.
    
    In order to provide sufficient data to train the DPM to distinguish collisions from non-collisions in the simulated environment, we used a selection of four basic traffic scenarios involving two vehicles approaching an intersection of orthogonal roadways.  Each of the model architectures/configurations were trained and tested using examples from all four scenarios.  In the first scenario, the second vehicle approached the intersection from the sensor vehicle's right side, as shown in Figure \ref{fig:road-sensor-bar}; in the second scenario, the second vehicle approached from the left.  In the third scenario, the second vehicle approached the intersection from the opposite direction as the sensor vehicle, but in a different lane (head-on miss).  In the fourth scenario, the second vehicle approached from the opposite direction as the sensor vehicle, but in the same lane (head-on collision).  In each scenario, both vehicles were initially stationary, but in the first time step the second vehicle was commanded to immediately accelerate to top speed at maximum torque.  By varying the time delay before the sensor vehicle was commanded to begin moving, we could control the class of the training sample (collision or no-collision) for the first two scenarios.  For the third scenario, the class was always ``no-collision'' and for the fourth scenario, the class was always ``collision,'' but we used a variable delay for the sensor vehicle accelerator position anyway, to allow additional variation in the collected data.
    
    In all scenarios, no human involvement was required to label the outcomes.  Using the recorded positions of the sensor vehicle and second vehicle, along with collision-detection features of the simulation platform, the label of each training sample was determined automatically.  Further processing of the recorded scenario data was used to determine the time of the collision (or time of closest approach, for no-collision scenarios).  Using this time, the collected data for each training scenario was truncated to include only those samples within five seconds or less prior to the time of collision (or closest approach).  The resulting data was divided into overlapping five-frame training examples.  
    
    Training and validation of the deep predictive model proceeded in batches, for a fixed number of training iterations (unless learning during training indicated further training would be futile, in which case `early-stopping' was used to cut off further training).  The resulting trained model was tested using sequence data collected during the same simulation episodes as the training or validation, but not used for either purpose.  K-fold cross-validation was used with k=10\cite{kohavi1995study}.

\subsection{Results}

The following sections describe the experimental results for different settings of network architectures and model parameters. 

\subsubsection{Prediction using Multiple Cameras}
\label{sec:mult}
    Using a DPM with all three camera images as input and training with data from multiple samples from each of the dynamic street scenes, the deep predictive model learned to recognize and distinguish impending collisions from no-collisions.  Referring to Table \ref{fig:isa-sanityonly}, the deep predictive model which took as input sensor images + state data + action data achieved Matthews Correlation Coefficient (MCC)\cite{powers2011evaluation} values above 0.6 for testing data that closely followed the training scenarios. Regarding generalization testing, the DPM configuration which took as input sensor images only (`images-only' or `all-three'), achieved accuracy values above 0.68 and MCC values above 0.35 for testing data that did not resemble the training data.  Regarding variation of the DPM configuration for model input/architecture, however, the results for the three configurations varying model input type are not statistically distinct, as discussed below. 

	Figure \ref{fig:road-sensor-bar} illustrates the output of the DPM for the first scenario, in which traffic approaches from the right side, including results for 1000 stochastic forward passes (SFPs).  The upper part of Figure \ref{fig:road-sensor-bar} shows the probability of collision is approximately equal to the probability of no-collision with a sizable variance, reflecting the uncertainty of the model when the vehicles are relatively far apart.  In contrast, the lower part of Figure \ref{fig:road-sensor-bar} shows the probability of collision is relatively smaller and the attendant histogram for 1000 SFPs exhibits a reduced variance, reflecting that the model is more certain about its prediction that no collision is likely, e.g., since the sensor images indicate the traffic will pass through the intersection without colliding. 
    
\begin{figure*}
    \centering
    \includegraphics[width=0.99\textwidth]{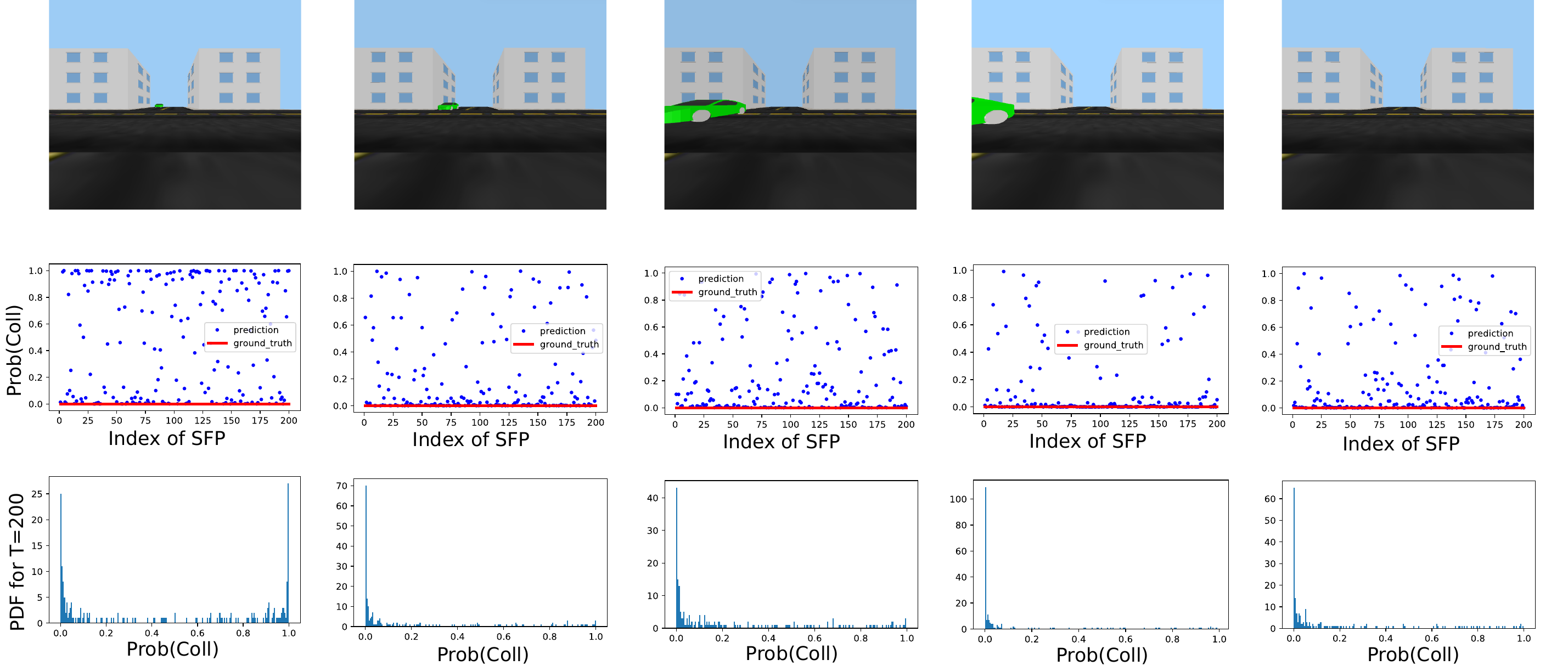}
    \caption{Histograms with varying distribution for no-collision scenario.}
    \label{fig:five-images}
\end{figure*}

    Figure \ref{fig:five-images} shows the output of the DPM for examples taken from the third scenario, in which traffic approaches from a head-on position, but in the opposite lane.  For five different times during the scenario, Figure \ref{fig:five-images} shows a sensor image for the first time step in the sequence, the DPM output for 200 stochastic forward passes (SFPs), and the same data presented as a frequency histogram.  Initially, the DPM output for 200 SFPs approximates a bimodal distribution, then reverts to something approaching a normal distribution as the traffic vehicle passes without colliding. This is an interesting insight, since it \textbf{suggests that different types of uncertainty may be present}. The network may be uncertain about a single possible outcome, i.e., one peak with high variance, or may be completely conflicting about the future, i.e., bi-modal distribution with probability mass on both sides of the spectrum.  

%
%

\subsubsection{Varying the Network Architecture}
Next, we performed experiments with different setups for the architecture of the DPM. To understand the influence of the proprioceptive data, as well as the action data, we performed an ablation study in which these input types where disabled. The results for varying the types of data input to the DPM are shown in Table \ref{fig:varyinput-sanityonly}, including a comparison of three configurations: \emph{images-only, images+state}, and \emph{images+state+action}. The first configuration employed a model architecture having three input images, one each from the left mirror, dashcam, and right mirror, and without state or action data inputs to the model.  As a result, the model architecture employed three branches of the type shown in the upper (convolutional recurrent) branch of Figure \ref{fig:arch}, and did not employ the lower (recurrent) input branch shown. The DPM architecture for the second and third configurations, however, employed four branches:  the three branches described above for the images-only configuration, plus the recurrent (LSTM) branch shown accepting state and action data.  For the second configuration (images+state), state information was fed to the DPM.  For the third configuration (images+state+action), state and action information were fed to the DPM.  

%
%
\begin{table}
\renewcommand{\arraystretch}{1.3}
\caption{Comparison of K-fold Cross Validation\cite{kohavi1995study} results for varying model input/architecture (mean +/- std. dev.)}
\label{fig:varyinput-sanityonly}
\centering
	\begin{tabular}{l|*{3}{c}}
	\hline
		Model Input/Arch. & accuracy & MCC\cite{powers2011evaluation} \\
	\hline
	\hline
        Images-only & 0.8195 (+/- 0.045) & 0.6421 (+/- 0.089) \\
        Images+State & 0.7701 (+/- 0.045) & 0.5581 (+/- 0.084) \\
        Images+State+Action & 0.8219 (+/- 0.079) & 0.6484 (+/- 0.152) \\
	\hline
	\hline
	\end{tabular}
\end{table}

As Table \ref{fig:varyinput-sanityonly} shows, the `images+state+action' configuration showed the best classification accuracy, correctly predicting collisions and misses for over eighty percent of the test examples, resulting in an MCC of over 0.6.  However, the other configurations showed similar prediction accuracy.  This surprising result indicates that augmenting the DPM architecture to accept additional input data indicating vehicle speed, location, etc. did not immediately translate into significantly improved prediction accuracy. This indicates that \textbf{naive augmentation strategies for state-action conditioning, may need to be revisited and improved}.    

To verify that these results represented distinct groupings, Analysis of variance (ANOVA) tests was performed on all groups of data.  The one-way F-test on the accuracy results for these three groups showed an F-value of 2.238 and a p-value of 0.126.  For the MCC results, the one-way F-test showed an F-value of 1.799 and a p-value of 0.185.  These results, in particular the p-value above 0.05, indicate the three groups are not statistically distinct.  

%
%
\begin{table}
\caption{K-fold Cross Validation results for the DPM with images + states + action inputs on dataset similar to training dataset (train on 4500 samples, test on 384 of 500 samples in k-fold)}
\label{fig:isa-sanityonly}
\centering
	\begin{tabular}{l|*{3}{c}}
	\hline
		Config & accuracy & MCC \\
	\hline
	\hline
		k=1 (i+s+a) & 0.8099 & 0.6218 \\
		k=2 (i+s+a) & 0.8646 & 0.7326 \\
        k=3 (i+s+a) & 0.8125 & 0.6275 \\
        k=4 (i+s+a) & 0.8854 & 0.7704 \\
        k=5 (i+s+a) & 0.9427 & 0.8817 \\
        k=6 (i+s+a) & 0.7031 & 0.4435 \\
        k=7 (i+s+a) & 0.7891 & 0.5787 \\
        k=8 (i+s+a) & 0.8307 & 0.6672 \\
        k=9 (i+s+a) & 0.6797 & 0.3597 \\
        k=10 (i+s+a) & 0.9010 & 0.8012 \\
        Mean (i+s+a) & 0.8219 & 0.6484 \\
        Std. Dev. (i+s+a) & 0.0790 & 0.1521 \\
	\hline
	\hline
	\end{tabular}
\end{table}

In an attempt to better understand the data, the accuracy and MCC results for each k-fold in K-fold cross validation were analyzed.  See, e.g., Table \ref{fig:isa-sanityonly}.  The logical conclusion is that the results reported in Table \ref{fig:varyinput-sanityonly} are unreliable for the purpose of drawing conclusions based on the differences in results between groups.  

\subsubsection{Varying Camera Perspective}
The results for varying camera position are shown in Table \ref{fig:varycam-sanityonly}, including a comparison of four configurations: Left mirror, Dashcam, Right mirror, and all three of the above.  The first three configurations employed the same model architecture having a single image input, and without state or action data inputs to the model.  As a result, the model architecture employed a single branch (i.e., the top branch shown in Figure \ref{fig:arch}.  The DPM for the Left mirror configuration was fed image sequences sourced from an image sensor located at the left-hand rear-view mirror on the simulated vehicle, angled 45 degrees to the left of the vehicle direction.  Likewise, the DPM for the Right mirror configuration was fed images sequences sourced from an image sensor located at the right-hand rear-view mirror on the simulated vehicle, angled 45 degrees to the right of the vehicle direction.  The Dashcam configuration was fed images from an image sensor located where a dashcam is most commonly located on the vehicle, near the center of the windshield.  Finally, the fourth configuration in this comparison was fed image data from all three of the aforementioned image sensor locations,  requiring a change to the model architecture to process three images every frame, rather than one.  As a result, the DPM architecture employed three branches, each using convolutional-recurrent nodes as shown in the top branch in Figure \ref{fig:arch}.  Thus, this `all three' configuration was identical to the `images-only' configuration from the model input study.

As Table \ref{fig:varycam-sanityonly} shows, \textbf{the `all three' configuration showed the best classification accuracy}, correctly predicting collisions and misses for over eighty percent of the test examples, resulting in a Matthews Correlation Coefficient (MCC) of over 0.6.  The other configurations showed lower prediction accuracy.  One surprising result was that the Right mirror configuration showed a lower level of accuracy than the Left mirror configuration.  This result may be explained by considering that, of the four training scenarios, two favored visibility on the left side of the vehicle.  The second (traffic from left) and third (head-on miss) scenarios each involved the second car approaching the sensor car from its left side, while only the first scenario (traffic from right) favored the Right mirror configuration.  On the other hand, the Dashcam configuration also showed a lower level of accuracy despite two of the four configurations involving the second car approaching the sensor from the front.  

To verify that the results in Table \ref{fig:varycam-sanityonly} resulted from statistically distinct groupings, Analysis of variance (ANOVA) tests were performed on the four groups of data.  The one-way F-test on the accuracy results for these four groups showed an F-value of 8.039 and a p-value of 0.0003.  For the MCC results, the one-way F-test showed an F-value of 8.262 and a p-value of 0.0003.  These results are all within conventional norms indicating the four groups were statistically distinct.

\section{Discussion}

The above experiments revealed interesting results regarding the application of Deep Predictive Models for collision risk assessment. A first important insight is that ``more is better'' in this domain: using multiple cameras as input to the DPM produced the best prediction rates without requiring any changes to the underlying machine learning framework. In contrast, the second study revealed that using proprioceptive data and action commands may not immediately lead to any improvement in prediction accuracy. This is a surprising result and may indicate that more sophisticated methods for incorporating state-action conditioning are needed. Another critical insight from the above experiments is that uncertainty information generated by the Bayesian ConvLSTM provides information about three classes of predictions: (a) high certainty predictions with low variance in the network output, (b) low certainty predictions with high variance in the network output, and interestingly also (c) conflicting predictions with bimodal output of the network. Understanding the difference between these three situations may yield important conclusions about \emph{when} and \emph{how} to use DPMs. In traditional Deep Learning methods, e.g., methods that generate only a point estimate, there is no possibility to distinguish between these situations.      

%
%
\begin{table}
\renewcommand{\arraystretch}{1.3}
\caption{Comparison of K-fold Cross Validation results for varying camera position (mean +/- std. dev.)}
\label{fig:varycam-sanityonly}
\centering
	\begin{tabular}{l|*{3}{c}}
	\hline
		Camera Position & accuracy & MCC \\
	\hline
	\hline
        L mirror & 0.7471 (+/- 0.075) & 0.5035 (+/- 0.140) \\
        Dashcam & 0.7089 (+/- 0.068) & 0.4311 (+/- 0.129) \\
        R mirror & 0.6870 (+/- 0.053) & 0.3875 (+/- 0.101) \\
        All 3 & 0.8195 (+/- 0.045) & 0.6421 (+/- 0.089) \\
	\hline
	\hline
	\end{tabular}
\end{table}

\section{CONCLUSIONS}
In this paper, we presented a novel methodology for assessing collision risk using Deep Predictive Models. In particular, we introduced a specific method called Bayesian ConvLSTMs for spatio-temporal processing of visual data, proprioceptive data and steering commands to identify potential impending collisions. In contrast to other approaches to Deep Learning in robotics, our approach allows for probabilistic beliefs over the output of the neural network. This information can, in turn, be used to assess the uncertainty inherent in the prediction.

A number of simulation experiments revealed important insights regarding the use of DPMs for collision risk assessment. In particular, our experiments indicate that processing multiple camera images simultaneously within the same network architecture is feasible and beneficial for this domain. While the prediction results are encouraging ($>80\%$), there is still room for improvement in this regard. For future work, we hope to investigate how to better leverage the available proprioceptive data.

\addtolength{\textheight}{-12cm}   

\section*{ACKNOWLEDGMENT}

This work was supported in part by the NSF I/UCRC Center for Embedded Systems (CES) and from NSF grants 1361926 and 1446730.  The authors would also like to thank Bosch and Toyota for their support and feedback through CES.

\bibliographystyle{./IEEEtran} 
\bibliography{./IEEEabrv,./IEEEexample}

\begin{thebibliography}{10}
\providecommand{\url}[1]{#1}
\csname url@rmstyle\endcsname
\providecommand{\newblock}{\relax}
\providecommand{\bibinfo}[2]{#2}
\providecommand\BIBentrySTDinterwordspacing{\spaceskip=0pt\relax}
\providecommand\BIBentryALTinterwordstretchfactor{4}
\providecommand\BIBentryALTinterwordspacing{\spaceskip=\fontdimen2\font plus
\BIBentryALTinterwordstretchfactor\fontdimen3\font minus
  \fontdimen4\font\relax}
\providecommand\BIBforeignlanguage[2]{{%
\expandafter\ifx\csname l@#1\endcsname\relax
\typeout{** WARNING: IEEEtran.bst: No hyphenation pattern has been}%
\typeout{** loaded for the language `#1'. Using the pattern for}%
\typeout{** the default language instead.}%
\else
\language=\csname l@#1\endcsname
\fi
#2}}

\bibitem{Deepdrive}
\BIBentryALTinterwordspacing
U.~of~Calif.~Regents. (2017) Berkeley deepdrive. [Online]. Available:
  \url{http://www.path.berkeley.edu/berkeley-deepdrive}
\BIBentrySTDinterwordspacing

\bibitem{chen2017rear}
C.~Chen, H.~Xiang, T.~Qiu, C.~Wang, Y.~Zhou, and V.~Chang, ``Rear-end collision
  prediction scheme based on deep learning in the internet of vehicles,''
  \emph{Journal of Parallel and Distributed Computing}, 2017.

\bibitem{DBLP:journals/corr/HamletC15}
\BIBentryALTinterwordspacing
A.~J. Hamlet and C.~D.~C. III, ``Joint belief and intent prediction for
  collision avoidance in autonomous vehicles,'' \emph{CoRR}, vol.
  abs/1504.00060, 2015. [Online]. Available:
  \url{http://arxiv.org/abs/1504.00060}
\BIBentrySTDinterwordspacing

\bibitem{Powar}
G.~Powar, ``{Calculation of collision probability for autonomous vehicles using
  trajectory prediction},'' Master's thesis, Rutgers University, New Brunswick,
  2016.

\bibitem{chang2010intelligent}
B.~R. Chang, H.~F. Tsai, and C.-P. Young, ``Intelligent data fusion system for
  predicting vehicle collision warning using vision/gps sensing,'' \emph{Expert
  Systems with Applications}, vol.~37, no.~3, pp. 2439--2450, 2010.

\bibitem{DBLP:journals/corr/abs-1202-2160}
\BIBentryALTinterwordspacing
C.~Farabet, C.~Couprie, L.~Najman, and Y.~LeCun, ``Scene parsing with
  multiscale feature learning, purity trees, and optimal covers,'' \emph{CoRR},
  vol. abs/1202.2160, 2012. [Online]. Available:
  \url{http://arxiv.org/abs/1202.2160}
\BIBentrySTDinterwordspacing

\bibitem{DBLP:journals/corr/DrewsWGTR17}
\BIBentryALTinterwordspacing
P.~Drews, G.~Williams, B.~Goldfain, E.~A. Theodorou, and J.~M. Rehg,
  ``Aggressive deep driving: Model predictive control with a {CNN} cost
  model,'' \emph{CoRR}, vol. abs/1707.05303, 2017. [Online]. Available:
  \url{http://arxiv.org/abs/1707.05303}
\BIBentrySTDinterwordspacing

\bibitem{hadsell2009learning}
R.~Hadsell, P.~Sermanet, J.~Ben, A.~Erkan, M.~Scoffier, K.~Kavukcuoglu,
  U.~Muller, and Y.~LeCun, ``Learning long-range vision for autonomous off-road
  driving,'' \emph{Journal of Field Robotics}, vol.~26, no.~2, pp. 120--144,
  2009.

\bibitem{DBLP:journals/corr/HuvalWTKSPARMCM15}
\BIBentryALTinterwordspacing
B.~Huval, T.~Wang, S.~Tandon, J.~Kiske, W.~Song, J.~Pazhayampallil,
  M.~Andriluka, P.~Rajpurkar, T.~Migimatsu, R.~Cheng{-}Yue, F.~Mujica,
  A.~Coates, and A.~Y. Ng, ``An empirical evaluation of deep learning on
  highway driving,'' \emph{CoRR}, vol. abs/1504.01716, 2015. [Online].
  Available: \url{http://arxiv.org/abs/1504.01716}
\BIBentrySTDinterwordspacing

\bibitem{cho2014multi}
H.~Cho, Y.-W. Seo, B.~V. Kumar, and R.~R. Rajkumar, ``A multi-sensor fusion
  system for moving object detection and tracking in urban driving
  environments,'' in \emph{Robotics and Automation (ICRA), 2014 IEEE
  International Conference on}.\hskip 1em plus 0.5em minus 0.4em\relax IEEE,
  2014, pp. 1836--1843.

\bibitem{DBLP:journals/corr/JainKSRSS16}
\BIBentryALTinterwordspacing
A.~Jain, H.~S. Koppula, S.~Soh, B.~Raghavan, A.~Singh, and A.~Saxena,
  ``Brain4cars: Car that knows before you do via sensory-fusion deep learning
  architecture,'' \emph{CoRR}, vol. abs/1601.00740, 2016. [Online]. Available:
  \url{http://arxiv.org/abs/1601.00740}
\BIBentrySTDinterwordspacing

\bibitem{DBLP:journals/corr/KendallBC15}
\BIBentryALTinterwordspacing
A.~Kendall, V.~Badrinarayanan, and R.~Cipolla, ``Bayesian segnet: Model
  uncertainty in deep convolutional encoder-decoder architectures for scene
  understanding,'' \emph{CoRR}, vol. abs/1511.02680, 2015. [Online]. Available:
  \url{http://arxiv.org/abs/1511.02680}
\BIBentrySTDinterwordspacing

\bibitem{kendall2016modelling}
A.~Kendall and R.~Cipolla, ``Modelling uncertainty in deep learning for camera
  relocalization,'' in \emph{Robotics and Automation (ICRA), 2016 IEEE
  International Conference on}.\hskip 1em plus 0.5em minus 0.4em\relax IEEE,
  2016, pp. 4762--4769.

\bibitem{Tesla}
\BIBentryALTinterwordspacing
F.~Lambert. (2017) Tesla has opened the floodgates of autopilot data gathering.
  [Online]. Available:
  \url{https://electrek.co/2017/06/14/tesla-autopilot-data-floodgates/}
\BIBentrySTDinterwordspacing

\bibitem{Mtyai}
\BIBentryALTinterwordspacing
M.~Shobe. (2017) Sample training dataset for autonomous driving now available.
  [Online]. Available:
  \url{https://mty.ai/blog/new-open-training-dataset-for-autonomous-driving-now-available/}
\BIBentrySTDinterwordspacing

\bibitem{Cityscapes}
\BIBentryALTinterwordspacing
M.~Cordts. (2017) Cityscapes dataset. [Online]. Available:
  \url{https://www.cityscapes-dataset.com/}
\BIBentrySTDinterwordspacing

\bibitem{Geiger2013IJRR}
A.~Geiger, P.~Lenz, C.~Stiller, and R.~Urtasun, ``Vision meets robotics: The
  kitti dataset,'' \emph{International Journal of Robotics Research (IJRR)},
  2013.

\bibitem{Udacity}
\BIBentryALTinterwordspacing
O.~Cameron. (2016) Open sourcing 223gb of driving data collected in mountain
  view, ca by our lincoln mkz. [Online]. Available:
  \url{https://medium.com/udacity/open-sourcing-223gb-of-mountain-view-driving-data-f6b5593fbfa5}
\BIBentrySTDinterwordspacing

\bibitem{hochreiter1997long}
S.~Hochreiter and J.~Schmidhuber, ``Long short-term memory,'' \emph{Neural
  computation}, vol.~9, no.~8, pp. 1735--1780, 1997.

\bibitem{DBLP:journals/corr/Graves13}
\BIBentryALTinterwordspacing
A.~Graves, ``Generating sequences with recurrent neural networks,''
  \emph{CoRR}, vol. abs/1308.0850, 2013. [Online]. Available:
  \url{http://arxiv.org/abs/1308.0850}
\BIBentrySTDinterwordspacing

\bibitem{DBLP:journals/corr/ShiCWYWW15}
\BIBentryALTinterwordspacing
X.~Shi, Z.~Chen, H.~Wang, D.~Yeung, W.~Wong, and W.~Woo, ``Convolutional {LSTM}
  network: {A} machine learning approach for precipitation nowcasting,''
  \emph{CoRR}, vol. abs/1506.04214, 2015. [Online]. Available:
  \url{http://arxiv.org/abs/1506.04214}
\BIBentrySTDinterwordspacing

\bibitem{Gal2016Uncertainty}
Y.~Gal, ``Uncertainty in deep learning,'' Ph.D. dissertation, University of
  Cambridge, 2016.

\bibitem{vrep2013}
M.~F. E.~Rohmer, S. P. N.~Singh, ``V-rep: a versatile and scalable robot
  simulation framework,'' in \emph{Proc. of The International Conference on
  Intelligent Robots and Systems (IROS)}, 2013.

\bibitem{kohavi1995study}
R.~Kohavi \emph{et~al.}, ``A study of cross-validation and bootstrap for
  accuracy estimation and model selection,'' in \emph{Ijcai}, vol.~14,
  no.~2.\hskip 1em plus 0.5em minus 0.4em\relax Stanford, CA, 1995, pp.
  1137--1145.

\bibitem{powers2011evaluation}
D.~M.~W. Powers, ``Evaluation: from precision, recall and f-measure to roc,
  informedness, markedness and correlation,'' \emph{International Journal of
  Machine Learning Technology}, vol.~2, no.~1, pp. 37--63, 2011.

\end{thebibliography}

\end{document}